\documentclass[lettersize,journal]{IEEEtran}
\usepackage{amsmath,amsfonts}
\usepackage{algorithmic}
\usepackage{algorithm}
\usepackage{array}
\usepackage[caption=false,font=normalsize,labelfont=sf,textfont=sf]{subfig}
\usepackage{textcomp}
\usepackage{stfloats}
\usepackage{url}
\usepackage{verbatim}
\usepackage{graphicx}
\usepackage{cite}
\usepackage{booktabs}   
\usepackage{multirow}   
\usepackage{array}      
\usepackage{pifont}
\usepackage{multirow}
\usepackage{array}
\usepackage{xcolor}
\usepackage{graphicx}
\definecolor{mypink}{RGB}{255,105,180}
\usepackage[colorlinks=true, linkcolor=mypink, citecolor=mypink, urlcolor=mypink]{hyperref}
\usepackage{tabularx}
\usepackage{array}
\usepackage[table]{xcolor}  
\usepackage{booktabs}
\usepackage{amsmath}
\makeatletter
\def\@cite#1#2{{\hypersetup{citecolor=blue}\textcolor{blue}{[{#1\if@tempswa , #2\fi}]}}}
\begin{document}

\title{HA2F: Dual-module Collaboration-Guided \textbf{H}ierarchical \textbf{A}daptive \textbf{A}ggregation \textbf{F}ramework for Remote Sensing Change Detection}

\author{Shuying Li, Yuchen Wang, San Zhang, Chuang Yang, \IEEEmembership{Member,~IEEE,}
\thanks{
Shuying Li, Yuchen Wang and San Zhang are with the School of Artificial Intelligence and School of Automation, Xi’an University of Posts and Telecommunications, Xi’an 710121,China (e-mail: lishuying@xupt.edu.cn; wyc0116@stu.xupt.edu.cn; zhangsan@xupt.edu.cn).}
\thanks{Chuang Yang is with the Department of Electrical and Electronic Engineering, The Hong Kong Polytechnic University, Hong Kong SAR (e-mail:omtcyang@gmail.com).}
\thanks{Corresponding author: Chuang Yang.}}

\markboth{~}%
{Shell \MakeLowercase{\textit{et al.}}: Dual-module Collaboration Guided Hierarchical Adaptive Aggregation Framework for Remote Sensing Change Detection}

\maketitle

\begin{abstract}
Remote sensing change detection (RSCD) aims to identify the spatio-temporal changes of land cover, providing critical support for multi-disciplinary applications (e.g., environmental monitoring, disaster assessment, and climate change studies). Existing methods focus either on extracting features from localized patches, or pursue processing entire images holistically, which leads to the cross temporal feature matching deviation and exhibiting sensitivity to radiometric and geometric noise. Following the above issues, we propose a dual-module collaboration guided hierarchical adaptive aggregation framework, namely HA2F, which consists of dynamic hierarchical feature calibration module (DHFCM) and noise-adaptive feature refinement module (NAFRM). The former dynamically fuses adjacent-level features through perceptual feature selection, suppressing irrelevant discrepancies to address multi-temporal feature alignment deviations. The NAFRM utilizes the dual feature selection mechanism to highlight the change sensitive regions and generate spatial masks, suppressing the interference of irrelevant regions or shadows. Extensive experiments verify the effectiveness of the proposed HA2F, which achieves state-of-the-art performance on LEVIR-CD, WHU-CD, and SYSU-CD datasets, surpassing existing comparative methods in terms of both precision metrics and computational efficiency. In addition, ablation experiments show that DHFCM and NAFRM are effective. \href{https://huggingface.co/InPeerReview/RemoteSensingChangeDetection-RSCD.HA2F}{HA2F Official Code is Available Here!}
\end{abstract}


\section{Introduction}
\IEEEPARstart{R}{emote} sensing change detection (RSCD) is a key technology for monitoring land cover change, evaluating ecological environment evolution and planning urban sustainable development, which plays a vital role in achieving sustainable development and promoting social and economic growth. With the rapid development of remote sensing technology, high-resolution remote sensing images, with their rich spatial information, have become an indispensable data source for monitoring the dynamic changes of the earth's surface, providing unique advantages for accurately identifying surface changes. However, high resolution remote sensing images are easy to cause conflicts between large-scale land class changes and subtle structure changes, resulting in multi-scale difference feature matching deviation. For example, in urban environments with complex scale and shape, extreme changes in structural scale (from small wooden houses to skyscrapers) and irregular geometry make detection boundaries blurred and human targets confused~\cite{10900538}. At the same time, the enhanced spatial resolution image provides richer texture and morphological information, where vehicle movement, vegetation change, and other factors will lead to spurious changes~\cite{10965808}. In addition, under the influence of seasonal and illumination changes, buildings present inconsistent hue and texture changes due to different reflection properties~\cite{yang2025edge}. As a result, noise and spurious changes are generated.
\begin{figure}[!t]
	\centering
	\includegraphics[width=1\linewidth]{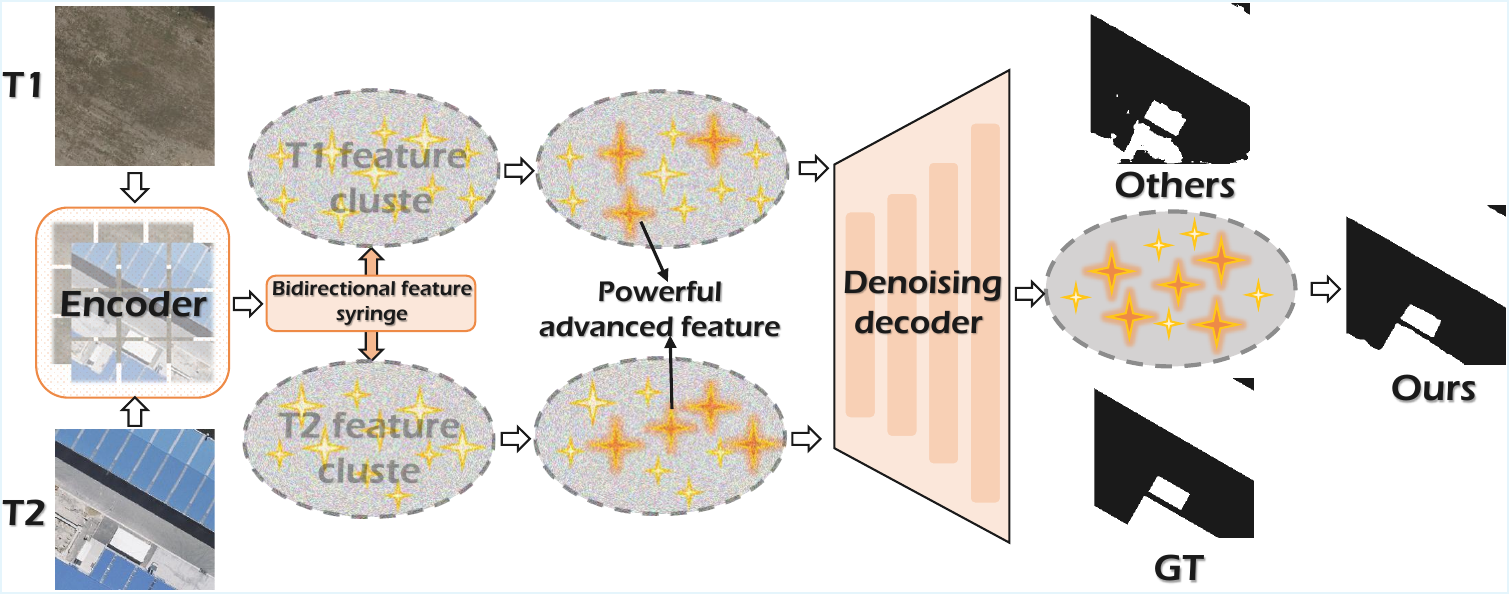} 
	\caption{Scheme of the motivation diagram. The proposed HA2F optimizes multi-scale difference feature fusion and purifies noise impurities.}
	\label{fig:0}
\end{figure}
To overcome the above issues, numerous methods have been proposed, which can be divided into Patch-based (PB) and Image-based (IB) methods roughly. PB-CD method extracts local blocks by sliding window or superpixel segmentation, and predicts the change state of pixels in an end-to-end training manner to complete the deep feature change detection~\cite{yang2025signeye}. Due to the scarcity of changing data in homogeneous scenarios, Gao~$et~al.$~\cite{gao2024building} proposed a metric guided supervised learning framework combined with a spatial attention module to generate pseudo-bi-temporal building change samples through patch pairing to suppress geometric offset and radiometric differences. On this basis, in order to further expand the applicability to heterogeneous data sources, Liu~$et~al.$~\cite{deng2024cross} input the patch sequence into the Transformer encoder and dynamically correlate multi-temporal features through cross-attention weights and the UperNet feature pyramid. To further improve efficiency, Xue~$et~al.$~\cite{10526384} rapidly filters a large number of unchanged image block pairs through sensitide-guided network pruning and multi-layer feature compression modules, significantly improving the processing efficiency on the edge computing platform. Although the hierarchical feature learning of patches can alleviate the limitation of feature matching at a single scale and improve the fusion effect of multi-scale difference features. However, the difference of sampling interval in different scale blocks will lead to feature misalignment in the edge regions, especially in high-frequency change regions such as building boundaries. The mismatch between the receptive fields of deep and shallow features may also cause the semantic gap. These multi-scale difference feature matching deviations will reduce the positioning accuracy of the change boundary.

Compared with PB-CD methods, IB-CD methods take dual temporal image segments as input, and learns the class change of each pixel through end-to-end training directly. The combination of Transformers and CNN has the significance in the current RSCD task. The IB-CD method based on CNN-Transformer hybrid combines the advantages of CNN's local feature extraction ability and Transformer's global context modeling, which significantly improves the change detection performance while reducing the model complexity. Yang~$et~al.$~\cite{10900538} designed a parallel convolution and multi-head self-attention mechanism interaction to integrate global and local information. Inspired by this, Badhe~$et~al.$~\cite{badhe2025hybrid} proposed a hybrid network that introduces a skeleton attention network to realize feature optimization, extracts spatio-temporal features with a three-branch CNN, and combines a parallel residual Transformer for global modeling. In order to reduce the computational complexity, Li~$et~al.$~\cite{10839538} designed a lightweight multi-scale feature fusion network by embedding a CNN structure in the Transformer backbone to enhance local feature extraction, which reduces the computational complexity and improves the representation ability of global-local features. Despite the excellent performance of IB-CD in end-to-end detection, it is still challenging to effectively integrate two heterogeneous architectures. It mainly stems from the local inductive bias of CNN, such as translation invariance, which is contradictory with the global attention mechanism of Transformer, leading to semantic misalignment and multi-spatio-temporal scale alignment difficulties during feature fusion~\cite{yang2023instance}. At the same time, IB-CD fails to fully model the physical laws and spatio-temporal context constraints of land surface changes in the feature extraction process, which is sensitive to noise and leads to spurious changes.

To this end, we propose a dual-module collaboration guided hierarchical adaptive aggregation (HA2F) framework, designed to aggregate multi-scale difference features, while suppressing the spurious changes caused by noise. Its motivation diagram is shown in Fig. \ref{fig:0}. In detail, The obtained bitemporal images are input into the hybrid architecture of Version Transformer (ViT) and ResNet18, through which global and local features are extracted respectively to realize the collaborative modeling of global context and local details. Subsequently, the multi-scale difference features are aggregated by the dynamic hierarchical feature calibration module (DHFCM). Most of the existing dynamic fusion modules in change detection rely on globally unified fusion weights or fixed attention patterns, which limits their sensitivity to spatially varying differences. In contrast, the proposed DHFCM introduces a hierarchical two-headed dynamic correction mechanism that performs finer and more adaptive feature fusion from both structural and global semantic levels. After extracting hierarchical difference features, a three-layer cross-attention mechanism is adopted, where low-level features serve as key and value vectors and high-level features serve as query vectors, ensuring the effective transfer of detailed spatial information into high-level semantics. The enhanced features are further processed by a bidirectional perceptual selection strategy, enabling cross-level temporal feature complementation and suppressing misalignment in multi-temporal representations. Afterwards, the difference enhancement features are input into the noise adaptive feature refinement module (NAFRM). Existing noise-aware refinement methods mostly depend on heuristic post-processing and lack learnable noise modeling capabilities. The proposed NAFRM differs fundamentally by learning pixel-wise noise bias estimation maps rather than relying on handcrafted thresholds or rules. The learned spatial bias field, generated through spatial-adaptive transformation, dynamically corrects multi-scale feature misalignment. Combined with a dual feature selection mechanism, a channel–space filtering strategy highlights change-sensitive regions in both low- and high-frequency features, while effectively suppressing illumination or shadow-induced spurious changes.

The main contributions of this work are summarized as follows:
\begin{enumerate}
\item The dynamic hierarchical feature calibration module (DHFCM) is designed to alleviate the problem of multi-scale difference feature fusion and structural information loss. It dynamically aggregates adjacent hierarchical features through content perception and maintains structural information, which helps enhance advanced features and mitigate uncorrelated differences.

\item The noise-adaptive feature refinement module (NAFRM) is desinged to optimize noise robustness in modal feature fusion. It uses the spatial bias field to achieve pixel-level non-rigid alignment, while employing the channel spatial filter to suppress the illumination or shadow noise interference in the irrelevant region and alleviate the spurious variation.

\item Based on the DHFCM and NAFRM modules, we construct a CNN-Transformer hybrid architecture (HA2F) for remote sensing change detection, which provides an efficient solution for the problems existing in current Patch-based (PB) and Image-based (IB) methods in multi-scale feature matching bias and noise sensitivity.
\end{enumerate}

The rest of this paper is structured as follows. Section \ref{sec:Related Work} overviews related works. Section \ref{sec:Methodology} describes the proposed HA2F framework in detail. The comprehensive experimental results and analysis are presented in Section \ref{Experimental Results and Analysis}. Finally, conclusions are drawn in Section \ref{CONCLUSION}. 

\section{Related Work}
\label{sec:Related Work}
\subsection{Patch-based (PB) methods} 

The utility of Patch-based methods is well-established in remote sensing image processing, where they serve as a vital tool for local modeling. In particular, with reference to the available ground-truth labels, Daudt~$et~al.$~\cite{daudt2018urban} for the first time proposed two convolutional neural network architectures based on Siamese and Early Fusion. Subsequently, this architecture became the foundation for many RSCD methods. Chen~$et~al.$~\cite{chen2020dasnet} designed Siamese ResUNet to enhance the discrimination of multi-scale Patch features through a dual attention mechanism and a weighted double margin contrastive loss function. Although this method could accurately distinguish real changes from noise, it had high computational complexity and is sensitive to sample balance. Zhou~$et~al.$~\cite{shangguan2023multi} modeled the spatial relationship between adjacent areas through the Patch graph structure, and combined the channel attention mechanism to enhance the ability to capture the details of complex ground objects. Graph structure modeling indirectly enhances the feature representation of sparsely varying regions by reasoning about vertex relationships, but it leads to blurred image edges. In order to reduce the local pixel-level uncertainty, Qing~$et~al.$~\cite{qing2022operational} proposed the pre-event superpixel constraint (PreSC), which solves the problem of superpixel edge blurring through multi-modal feature fusion and prior shape constraints. Although these methods have demonstrated their capabilities in RSCD, there is a significant bias in multi-scale matching. The fixed-size Patch segmentation limits the receptive field of features and cannot adapt to the change of object scale at different resolutions. At the same time, the hard boundary cut destroys the context continuity and leads to the failure of cross-scale feature association.
\begin{figure*}[!t]
	\centering
	\includegraphics[width=1\linewidth]{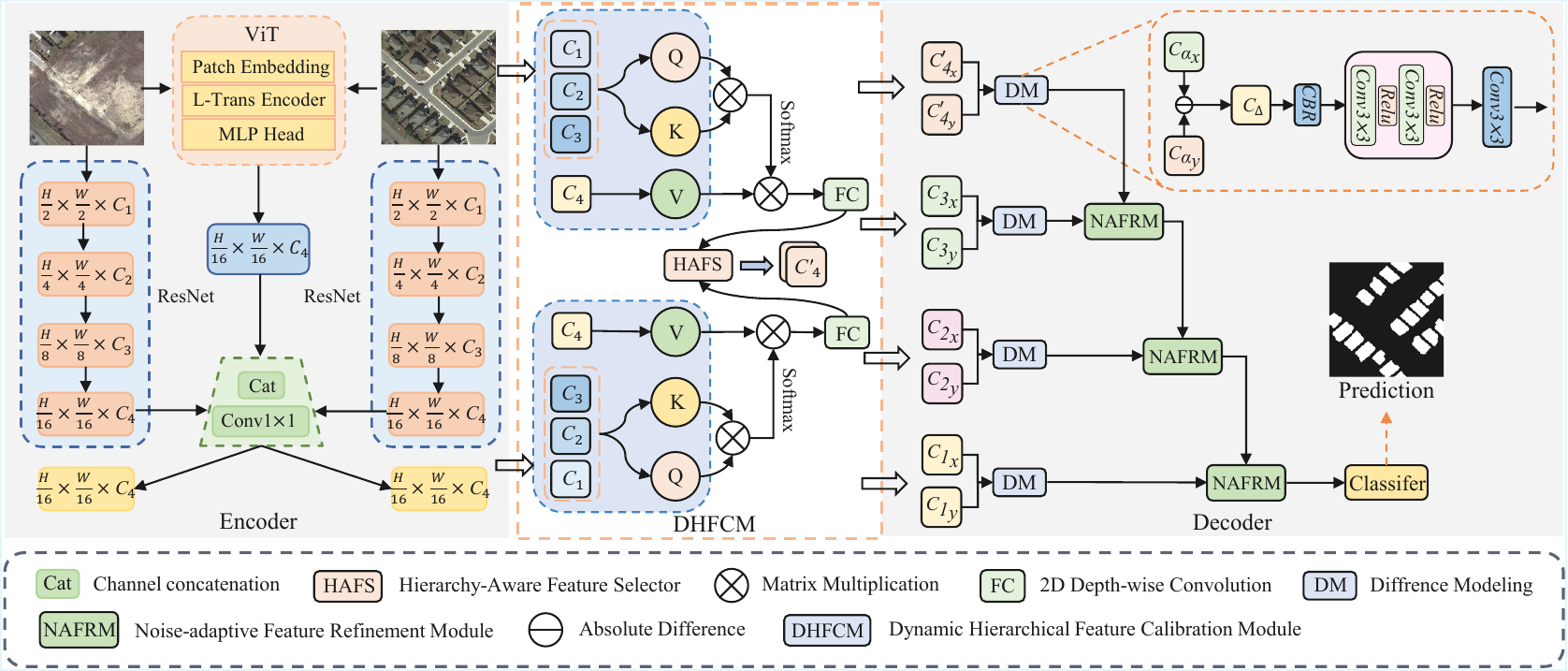} 
	\caption{Overview of the proposed HA2F. Bitemporal features are first extracted using ViT and ResNet18. The DHFCM then integrates cross-layer difference information to generate highly discriminative enhanced advanced features. Next, the NAFRM achieves feature alignment by spatially adaptive transformation and filters change sensitive regions to optimize noise robustness. Finally, the classifier generates a change graph that reflects the actual architecture changes.}
	\label{fig:1} 
\end{figure*}
\subsection{Image-based (IB) methods}   
With respect to the network architecture, IB-DLCD methods are classified into three sub-categories: convolutional neural network-based; Transformer-based; and hybrid CNN-Transformer frameworks. The preliminary RSCD network of CNN used the early fusion architecture to concatenate dual-temporal remote sensing images directly in the input layer. However, this approach was prone to lead to insufficient semantic information mining for a single time stage~\cite{han2024real} . To overcome this limitation, Ding~$et~al.$~\cite{ding2022bi} proposed a mid-term fusion framework. By introducing a dual semantic reasoning mechanism, the framework aimd to strengthen the semantic representation within a single time phase, and then modeled the semantic relevance across time phases. In CNN-based IB-DLCD, the species feature compression process at the encoder stage leads to severe fine-grained information loss, resulting in insufficient representation ability of fine edge structures. Compared with CNN network, Transformer model has powerful feature representation and context modeling capabilities. Zhang~$et~al.$~\cite{zhang2022swinsunet} designed the first transformer-based CD model, named SwinSUNet. The model introduces a feature fusion module, which effectively alleviates the problem of loss of detail information in the process of deep feature extraction. Although the feature exchange architecture of dual encoder-decoder achieves effective fusion of temporal features at the decision-making level, there are still limitations in terms of computational efficiency and detail preservation. For this reason, Noman~$et~al.$~\cite{noman2024remote} captured the inherent characteristics of the data by replacing the standard self-attention unit with a random sparse attention module to focus on regions with sparse changes. The IB-DLCD method based on the CNN-Transformer hybrid architecture innovatively empowers CNN and Transformer units into the feature extractor, and constructs an encoder that directly captures the global-local hybrid features. For instance, Li~$et~al.$~\cite{li2023convtransnet} proposed a parallel ConvTrans module for global and local feature extraction and fusion. However, the feature interaction mechanism still lacks explicit modeling of cross-temporal semantic relationships, resulting in limited ability to suppress spurious changes under complex geospatial distribution. Yang~$et~al.$~\cite{10900538} adopted the parallel structure of CNN and multi-head Self-Attention (MSA) to realize the efficient interaction between local features and global dependencies, enhanceing the adaptability of the model to complex scenes. This method reduces the computational complexity through window self-attention, but also lacks a more refined feature interaction mechanism. Therefore, it is urgent to explore an efficient multi-scale differential feature aggregation mechanism to suppress the accumulation of redundant features.

\section{Methodology}
\label{sec:Methodology}
\subsection{Architecture Overview} 
This section begins with an introduction to the overall architecture of our HA2F. Following this, detailed descriptions of each module in the network are provided.
                                                   
The overall network architecture of HA2F framework, based on an encoder-decoder structure, is illustrated in Fig. \ref{fig:1}. In the encoder stage, dual-temporal images are fed into the pre-trained ViT and ResNet18 backbones in parallel to produce multi-level features. The ViT extracts high-level features, $\mathbf{F}_{V}^{t} \in \mathbb{R}^{\frac{H}{16} \times \frac{W}{16} \times C_{4}}$, where $t \in \{1, 2\}$ represents two phases, while the Resnet18 acquires fine-grained multi-scale features $\mathbf{F}_{R_{i}}^{t} \in \mathbb{R}^{\frac{H}{2^{i}} \times \frac{W}{2^{i}} \times C_{i}} \quad (i \in \{1, 2, 3, 4\},\ t \in \{1, 2\})$ features extracted by Vit and ResNet18 are fused to obtain new advanced features $\mathbf{F}_{N}^{t} \in \mathbb{R}^{\frac{H}{16} \times \frac{W}{16} \times C'_{4}}$. Subsequently, DHFCM uses the three-layer cross-attention mechanism to propagate the extracted low-level features to high-level features, while hierarchical awareness feature selector (HAFS) generates highly discriminative enhanced advanced features by effectively integrating cross-layer difference information. In the decoder stage, NAFRM achieves feature alignment through the displacement information of spatial adaptive transformation and supimposes the change sensitive regions after the fusion of low and high-frequency features through DFSM. Finally, the classifier generates a change graph that reflects the real architectural changes.
\begin{figure*}[!t]
	\centering
	\includegraphics[width=0.8\linewidth]{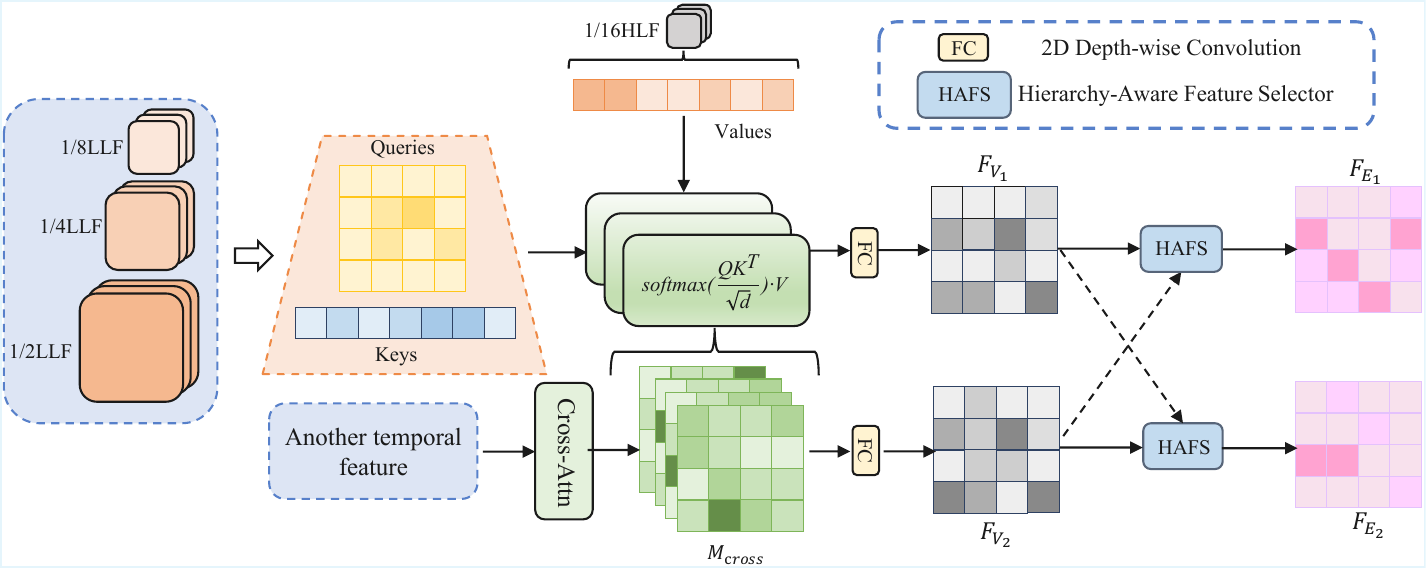} 
	\caption{Scheme of the DHFCM. In this module, the DHFCM mainly includes two parts: triple cross-attention mechanism and HAFS. Among them, the HLF represents high-level features and LLF represents low-level features.}
	\label{fig:2}
\end{figure*}
\begin{figure}[!t]
	\centering
	\includegraphics[width=1\linewidth]{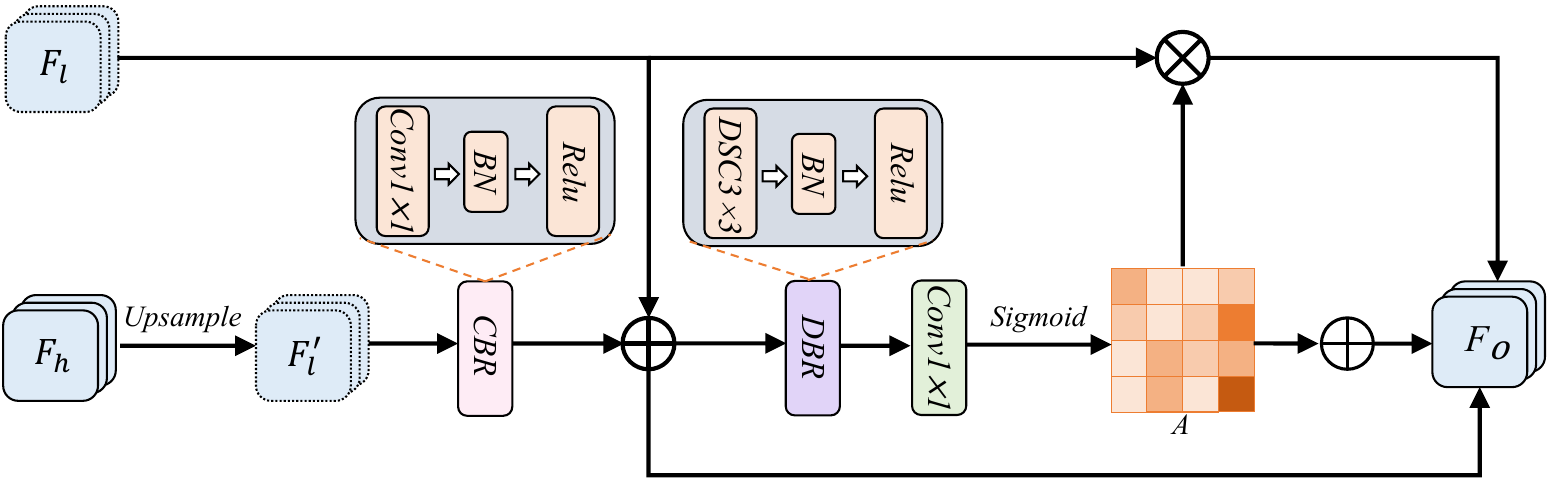} 
	\caption{Scheme of the HAFS in DHFCM.}
	\label{fig:3}
\end{figure}
\subsection{Feature Extraction Module}
In the encoder, the feature extraction module consists of a ViT and a 4-layer ResNet18 network. The bi-temporal images $T_{1}$ and $T_{2}$ are converted to a D-dimensional token sequence with position embedding (the space size $1/16$ of the original image). This sequence is then processed by a $L$ layer Transformer encoder, where each layer follows a sequence structure of: layer normalization (LN), multi-head self-attention (MHSA), residual connection, layer normalization, feedforward network (FFN), and residual connection for feature transformation\cite{zhu2024changevit}. The formulation of these layers can be expressed as:
\begin{equation}
	\label{deqn_ex1a}
	\mathbf{F}_{V}^{'t,\zeta+1} = \mathbf{F}_{V}^{t,\zeta} + \mathrm{MHSA}(\mathrm{LN}(\mathbf{F}_{V}^{t,\zeta})),
\end{equation}
\begin{equation}
	\label{deqn_ex2a}
	\mathbf{F}_{V}^{t,\zeta+1} = \mathbf{F}_{V}^{'t,\zeta+1} + \mathrm{FFN}(\mathrm{LN}(\mathbf{F}_{V}^{'t,\zeta+1})),
\end{equation}
where $\zeta$ denotes the output of the $\zeta$th transformer layer. The final output of the ViT backbone is denoted by $\mathbf{F}_{V}^{t} \in \mathbb{R}^{\frac{H}{16} \times \frac{W}{16} \times C_{4}}$ , where $C_{4}$ equals to $D$.

The ViT is good at capturing global semantics and large-scale changes with its self-attention mechanism. However, due to its structural characteristics of processing images in blocks, the ViT is simple to lose the detailed information of small objects, resulting in weak performance on subtle changes and small object detection tasks. Dealing with this challenges, we introduce the ResNet18 module for compensation. This module consists of four residual convolution blocks $(C_{1}-C_{4})$. After processing the input image through the 4-layer ResNet18 network, four scale detail features are generated, namely $1/2$, $1/4$, $1/8$ and $1/16$ representations, denoted as $\mathbf{F}_{R_{i}}^{t} \in \mathbb{R}^{\frac{H}{2^{i}} \times \frac{W}{2^{i}} \times C_{i}} \quad (i \in \{1, 2, 3, 4\})$. After concatenating the 1/16 feature map processed by ViT and the $1/16$ feature map processed by the 4-layer ResNet18 network in the channel dimension, the feature fusion is performed by $1\times1$ convolution to realize the collaborative modeling of global context and local details.

\begin{figure*}[!t]
	\centering
	\includegraphics[width=0.9\linewidth]{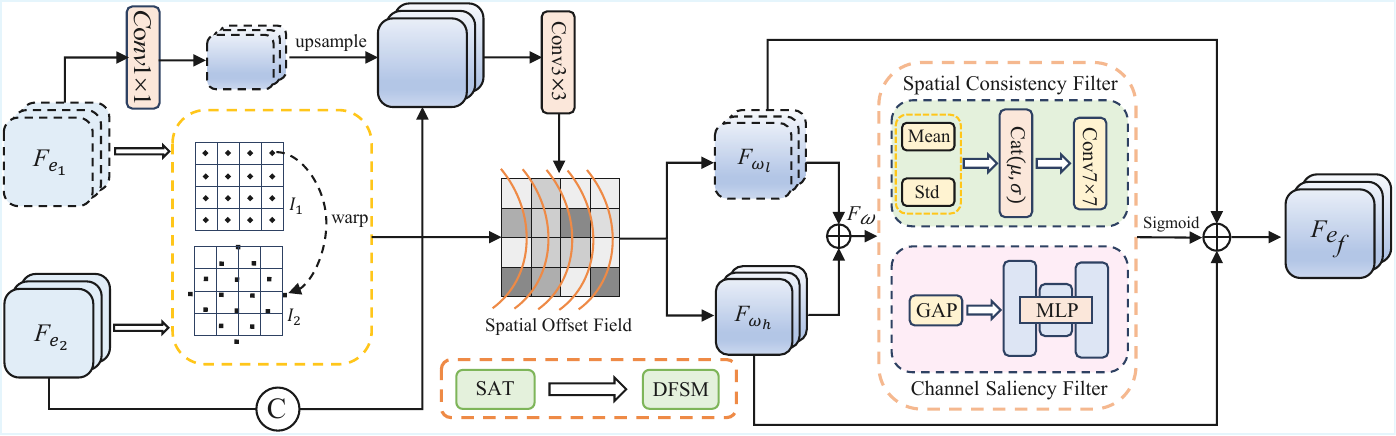} 
	\caption{Scheme of the NAFRM. In this module, SAT generates spatial bias field, after which HAFS filters noise information by double screening strategy.}
	\label{fig:4}
\end{figure*}

\subsection{Dynamic Hierarchical Feature Calibration Module}
In RSCD tasks, it is crucial to preserve detailed spatial information features, so it is necessary to ensure that low-level details effectively convey high-level semantic features. However, the simple upsampling in the passing process cannot recover the true detail information lost in the downsampling process. This will lead to the features cannot be accurately aligned in spatial position during fusion, and the edges of small objects will become blurred and inaccurate positioning. To this end, we introduce DHFCM, which consists of three cross-attention modules and HAFS. Their structures are shown in Fig. \ref{fig:2} and Fig. \ref{fig:3}.

The 3-layer cross-attention module uses the cross-layer feature propagation mechanism to integrate the low-level features of details into the high-level semantics of ViT. It treats the low-level features of $1/2$, $1/4$, $1/8$ as key and value vectors, and the high-level features of $1/16$ as query vectors, which are computed and output details are merged into the high-level representation of ViT, denoted as $\mathbf{F}_{V_{t}}^{i}$. The $\mathbf{F}_{V_{t}}^{i}$ can be computed as follows:
\begin{equation}
	\label{deqn_ex3a}
	\mathbf{F}_{V_{t}}^{i} = \mathrm{CrossAttn}(\mathbf{F}_{\frac{\mathit{HLF}}{16}}^{i}, \mathbf{F}_{\frac{\mathit{LLF}}{2^{i}}}^{i}),
\end{equation}
\begin{equation}
	\label{deqn_ex4a}
	\mathbf{F}_{V_{t}} = \mathrm{FC}(\mathbf{F}_{V_{t}}^{1} \copyright \mathbf{F}_{V_{t}}^{2} \copyright \mathbf{F}_{V_{t}}^{3}), 
\end{equation}
 where $i \in \{1, 2, 3\}$ represents the underlying index. The operation to $\mathbf{F}_{\frac{\mathit{HLF}}{16}}^{i}$ for a query, and respectively to $\mathbf{F}_{\frac{\mathit{LLF}}{2^{i}}}^{i}$ for key and value. Where HLF represents high-level features and LLF represents low-level features. The FC is a 2D depthwise separable convolution with kernel size $1\times1$ and $\copyright$ represents the concatenation of features along the channel dimension.
 
The cross-attention mechanism flattens 2D features into sequences, focusing on the semantic correlation between features and losing part of the spatial structure information. HAFS compensates for this by preserving spatial context information through convolution operations. At the same time, the bidirectional HAFS promotes feature consistency in non-changing regions, suppresses non-relevant differences, and alleviates multi-temporal feature alignment bias through mutual correction and attention reweighting. To clarify the convolutional units used in the proposed HAFS, we introduce the definitions of the CBR and DBR layers used throughout this section. The CBR block consists of a $1\times1$ convolution followed by batch normalization and a ReLU activation. It is used in HAFS to replace linear projections, enabling spatially aware feature transformation while preserving the 2D structure of feature maps. Then, The DBR block is composed of a $3\times3$ depthwise separable convolution followed by batch normalization and a ReLU activation. It enhances local spatial context modeling with significantly reduced parameters compared to standard convolutions. Finally, a spatial adaptive attention map of $H \times W \times 1$ dimensions is generated through the spatial attention mechanism, so that different locations can obtain different fusion weights according to their content importance. This process can be represented as:
\begin{equation}
	\label{deqn_ex5a}
	\mathbf{H}_{\mathrm{proj}} = \mathrm{ReLU}(\mathrm{BN}(\mathrm{Conv}_{1 \times 1}(\mathbf{F}_{l}))),
\end{equation}
\begin{equation}
	\label{deqn_ex6a}
	\mathbf{A} = \sigma(\mathrm{Conv}_{1 \times 1}(\mathrm{DBR}(\mathbf{F}_{l} + \mathbf{H}_{\mathrm{proj}}))),
\end{equation}
\begin{equation}
	\label{deqn_ex7a}
	\mathbf{F}_{o} = \mathbf{A} \odot \mathbf{F}_{l} + \mathbf{H}_{\mathrm{proj}},
\end{equation}
where $\mathbf{H}_{\mathrm{proj}}$ denotes feature projection. Additionally, $\sigma$ is the sigmoid activation function, and $\mathbf{A}$ denotes spatial attention map. Here, $\odot$  refers to element-wise multiplication.
\subsection{Noise-adaptive Feature Refinement Module}
In the decoder stage, the simple feature concatenation or addition operation will directly transfer the noise and irrelevant information to the subsequent layers, which is easy to produce spurious changes in the final output. To face this challenge, we introduce the NAFRM to perform geometrically fine alignment with noise adaptive filtering at the pixel level. This module consists of three stages: spatial bias field generation, forward warping and DFSM, as shown in fig. \ref{fig:4}.

In the spatial bias field generation stage, the $1\times1$ convolution is first used to adjust the channel number of the low-resolution feature map, so that it is consistent with the channel dimension of the high-resolution feature map. The spatial scale alignment is subsequently achieved by an upsampling operation, and a $3\times3$ convolution is used to generate a four-channel spatial bias field. The first two channels of the bias field encode the horizontal and vertical displacements of the high-resolution feature map, while the last two channels correspond to the displacements of the low-resolution feature map. The generation process can be expressed as follows:
\begin{equation}
	\label{deqn_ex8a}
	\mathbf{F}_{m} = \mathrm{Conv}_{3 \times 3}(\mathrm{Cat}(\mathbf{F}_{e_{2}}, \mathrm{up}(\mathrm{Conv}_{1 \times 1}(\mathbf{F}_{e_{1}})))),
\end{equation}
where $\mathbf{F}_{m}$ denotes the optical flow map, $\mathbf{F}_{e_{1}}$ represents the low-resolution feature map, and $\mathbf{F}_{e_{2}}$ represents the high-resolution feature map. $\mathrm{Conv}_{3 \times 3}$ and $\mathrm{Conv}_{1 \times 1}$ refer to the $3\times3$ and $1\times1$ convolutional operations, respectively.

In the forward warping phase, the position of each pixel in the feature map is remapped according to the displacement information provided by the bias field, and the bias calculation result is applied to the target image frame to obtain the transformed result image. If there are two adjacent images $I_{1}$ and $I_{2}$, it corresponds to the offset relationship $\mathbf{F}_{1 \rightarrow 2}$ of the corresponding points on the two images from $I_{1}$ to $I_{2}$. Under the forward warp operation, the pixel value of the first frame image ${I_{1}}(x,y)$ will appear at the coordinate position on the second frame image $I_{2}((x, y) + \mathbf{F}_{1 \rightarrow 2})$.
\begin{equation}
	\label{deqn_ex9a}
	I_{2}(x, y) = I_{1}(x + \Delta x, y + \Delta y),
\end{equation}
where ${I_{1}}(x,y)$ denotes the initial coordinates, while ${I_{2}}(x,y)$ signifies the transformation coordinates after forward warping. The $\mathbf{F}_{1 \rightarrow 2}(x, y)$ equals to $(\Delta x, \Delta y)$. Applying warping separately to the high-resolution and low-resolution feature maps contributes to more accurate alignment.

In the DFSM stage, the mechanism can adaptively suppress noise and enhance useful information through dual filtering of channel selection and space selection, so as to achieve more robust feature fusion. First, the warped feature maps $\mathbf{F}_{\omega_{l}}$ and $\mathbf{F}_{\omega_{h}}$ are added together to obtain the joint feature $\mathbf{F}_{\omega}$. Based on $\mathbf{F}_{\omega}$, spatial consistency filter is used to locate the key spatial regions in the feature map and suppress the background noise. The mean and standard deviation of each spatial location are calculated along the channel dimension, these two complementary statistics are concatenated, and fused through a $7\times7$ convolutional layer. Finally, the output is compressed to the 0-1 range by the sigmoid function to generate the spatial importance weight. At the same time, the channel saliency filter is used to suppress the redundancy or noise of the secondary channels with low weight value. After global average pooling (GAP) and multi-layer perceptron (MLP), the channel importance weights are generated by sigmoid function. The spatial and channel attention weights are then multiplied element-by-element, and the weighted feature maps are added to obtain the fusion feature map $\mathbf{F}_{e_{f}}$. This process can be represented as:
\begin{equation}
	\label{deqn_ex10a}
	\mathbf{F}_{\omega} = \mathbf{F}_{\omega_{l}} + \mathbf{F}_{\omega_{h}},
\end{equation}
\begin{equation}
	\label{deqn_ex11a}
	\omega_{s} = \sigma(\mathrm{MLP}(\mathrm{GAP}(\mathbf{F}_{\omega}))),
\end{equation}
\begin{equation}
	\label{deqn_ex12a}
	\omega_{c} = \sigma(\mathrm{Conv}_{7 \times 7}((\mu \oplus \varrho) \mathbf{F}_{\omega})),
\end{equation}
\begin{equation}
	\label{deqn_ex13a}
	\mathbf{F}_{e_{f}} = (\mathbf{F}_{\omega_{l}} + \mathbf{F}_{\omega_{h}}) \times \omega_{s} \times \omega_{c},
\end{equation}
where GAP represent global average pooling and $\oplus$ equals to channel-wise concatenation . The $\mu$ and $\varrho$ represent the channel mean and standard deviation, respectively.

\section{Experimental Results and Analysis}
\label{Experimental Results and Analysis}
\subsection{Datasets}
In the following experiments, three publicly available benchmark RSCD datasets are used and their synthesis details are outlined as follows.

1) \textbf{LEVIR-CD Dataset}~\cite{chen2020spatial}: This dataset comprises 637 meticulously annotated bi-temporal VHR image pairs (1024 $\times$ 1024 pixels, 0.5 m resolution) for RSCD, capturing building remodeling over 5-14 years. To optimize computational efficiency, we partitioned the images into non-overlapping 256 $\times$ 256 patches, resulting in standard splits of 7120 training, 1024 validation, and 2048 testing pairs.

2) \textbf{WHU-CD Dataset}~\cite{ji2018fully}: The dataset consists of a pair of optical images acquired in 2012 and 2016 of size 32507 $\times$ 15354 with the same spatial resolution of 0.2 m. Following the rules listed in the text, this dilated image is split into non-overlapping patches of pixel size 256 $\times$ 256. Then, these patches were divided into training, validation and test sets by random partitioning, resulting in 4536, 504, and 2760 pairs, respectively.

3) \textbf{SYSU-CD Dataset}~\cite{shi2021deeply}: The dataset consists of 20000 pairs of 256 $\times$ 256 pixels bi-temporal RSCD images, each with a spatial resolution of 0.5 m, taken between 2007 and 2014 in Hong Kong. It shows a variety of complex change scenarios such as road extension, new buildings, vegetation change, and suburban growth. The official 6:2:2 split divides all images in this dataset into 12000, 4000, and 4000 pairs for training, validation, and testing.
\subsection{Implementation Details and Evaluation Metrics}
\label{sec:Implementation Details and Evaluation Metrics}
1) \textbf{Implementation Details}: In our experiments, the proposed HA2F framework was implemented by the PyTorch platform and trained on an NVIDIA RTX 5090 GPU with 32GB of memory, employing a batch size of 16. We employ the Adam optimizer $(\beta_1=0.9, \beta_2=0.99)$ with a weight decay of 1e-4 for optimization. The initial learning rate is set to 2e-4 and follows a polynomial decay schedule with a power of 0.9 over 80K iterations. The pre-trained ResNet18 and ViT network models are used to extract features in the encoder, and the same partition mechanism as the comparison method is used for each dataset to ensure fairness. For data augmentation, we adopt random flipping and cropping to enhance model robustness. All comparative experiments are conducted under settings meticulously aligned with their original publications to ensure a fair comparison.

2) \textbf{Evaluation Metrics}: Following the widely used evaluation protocols in the change detection task, we use five accuracy metrics including precision (P), recall (R), overall accuracy (OA), F1 score (F1), and intersection over union (IoU) were used, which can provide a comprehensive assessment of the accuracy, completeness, and overall effectiveness of different RSCD methods. The above evaluation indicators are given as follows:
\begin{equation}
	\label{deqn_ex14a}
	\mathrm{P} =\mathrm{TP}/(\mathrm{TP}+\mathrm{FP}),
\end{equation}
\begin{equation}
	\label{deqn_ex15a}
	\mathrm{R} =\mathrm{TP}/(\mathrm{TP}+\mathrm{FN}),
\end{equation}
\begin{equation}
	\label{deqn_ex16a}
	\mathrm{OA} =\mathrm{TP+TN}/(\mathrm{TP+TN+FP+FN}),
\end{equation}
\begin{equation}
	\label{deqn_ex17a}
	\mathrm{F1} =\mathrm{2\times{P\times{R}}/(P+R)},
\end{equation}
\begin{equation}
	\label{deqn_ex18a}
	\mathrm{IoU} =\mathrm{TP/(TP+FP+FN)},
\end{equation}
where TP, TN, FP and FN denote the quantities of true positives, true negatives, false positives, and false negatives, respectively. The values of these indicators range from 0 to 1, with elevated values signifying superior performance.
\begin{figure*}[!t]
	\centering
	\includegraphics[width=1\linewidth]{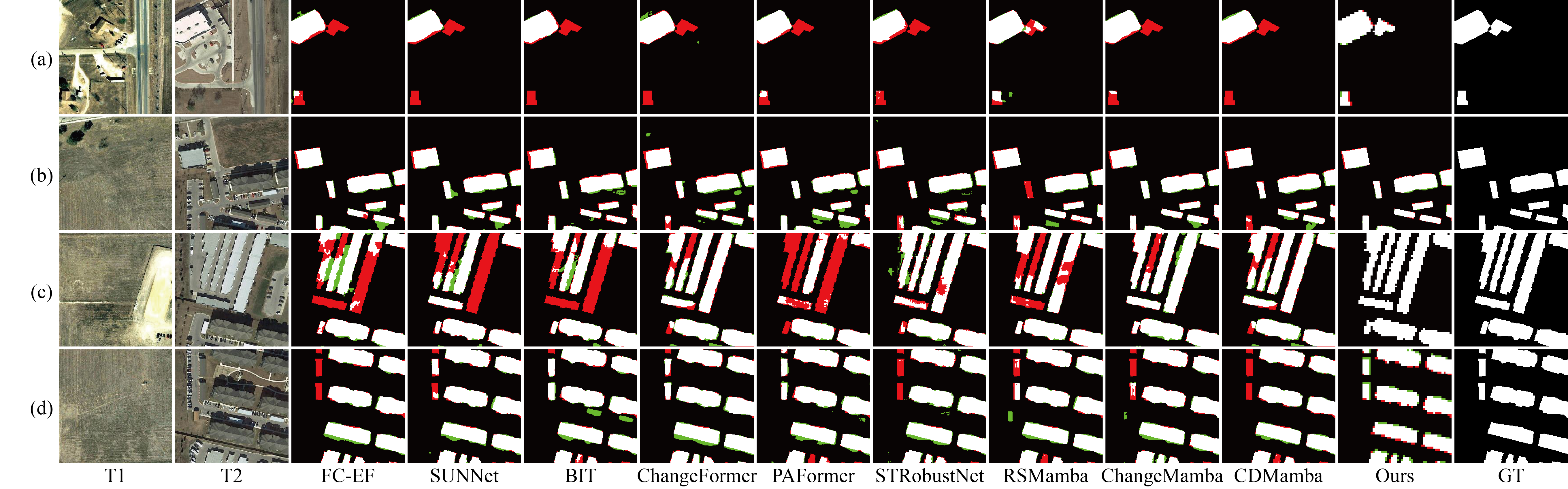} 
	\caption{Visual results of different RSCD methods on the LEVIR-CD dataset. (a)–(d) are four representative samples, where white and black represent changed and unchanged areas, while red and green indicate false detection pixels and missed detection pixels.}
	\label{fig:5}
\end{figure*}   
\begin{figure*}[!t]
	\centering
	\includegraphics[width=1\linewidth]{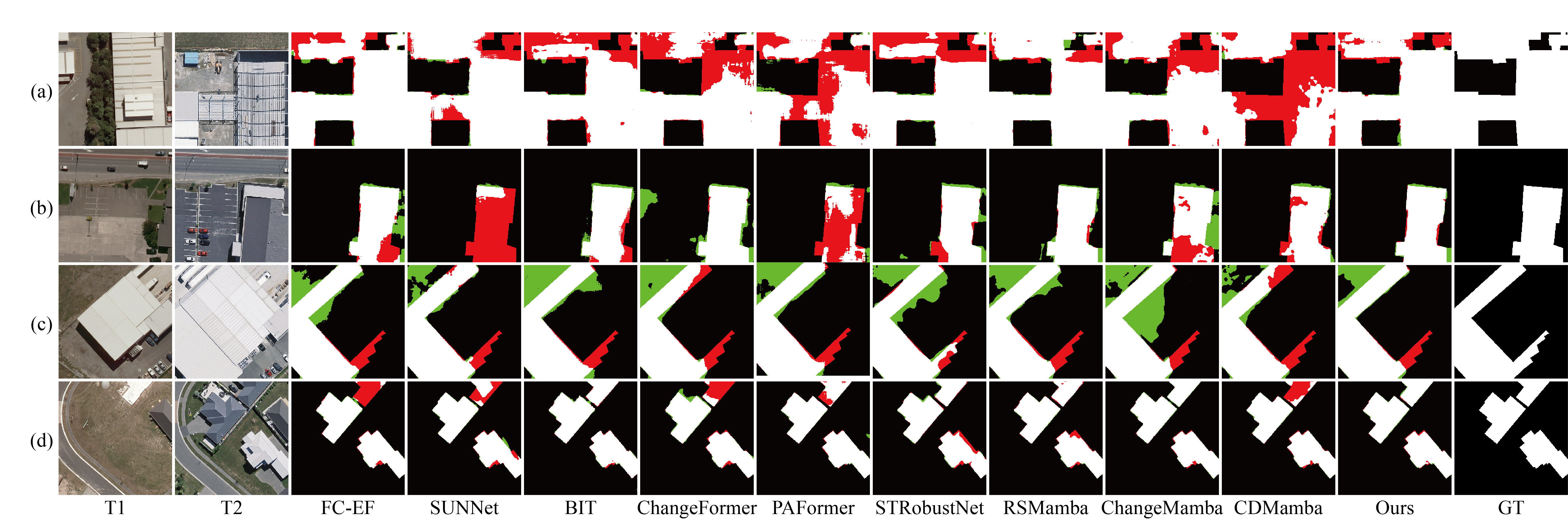} 
	\caption{Visual results of different RSCD methods on the WHU-CD dataset. (a)–(d) are four representative samples, where white and black represent changed and unchanged areas, while red and green indicate false detection pixels and missed detection pixels.}
	\label{fig:6}
\end{figure*}

\subsection{Comparative Methods}
To comprehensively evaluate the effectiveness of our network, we compare it with nine representative RSCD methods. To ensure fairness, the data augmentation techniques are also applied to all these models, and these related methods are summarized as follows:

1) \textbf{FC-EF}~\cite{daudt2018fully}: The study proposes two architectures based on siamese fully convolutional networks to detect remote sensing changes by introducing heuristic strategies.

2) \textbf{SNUNet}~\cite{9355573}: The study preserves high-resolution details with dense skip connections. The Siamese structure is used to ensure bitemporal relative alignment. Then, ECAM is used to fuse multi-level semantics to achieve accurate change detection in small targets, boundaries and complex scenes.

3) \textbf{BIT}~\cite{chen2021remote}: In this study, Transformer is used as the backbone network to improve the expression ability of fine-grained spatial differences and semantic consistency in remote sensing change detection by modeling long-distance inter-temporal dependencies and the interaction mechanism of dual-temporal features.

4) \textbf{ChangeFormer}~\cite{bandara2022transformer}: In this paper, the pure Transformer Siamese structure combines the learnable difference module and the lightweight decoder to realize the end-to-end global modeling. Compared with the BIT method, the hybrid structure of CNN+Transformer has more obvious advantages in particularly complex scenes.

5) \textbf{Paformer}~\cite{liu2022pa}: This study proposes prior feature extraction to obtain building structure priors, and uses a prior-aware Transformer to fuse these priors with bi-temporal deep features across time and space, thereby enhancing building contour understanding and long-distance dependency modeling simultaneously in an end-to-end framework.

6) \textbf{RSMamba}~\cite{10542538}: In this study, structured state-space models are used to replace the traditional CNN or transformer. Through the modeling ability of long sequences with linear complexity, the global features and spatial relationships of remote sensing images are efficiently extracted.

7) \textbf{ChangeMamba}~\cite{10565926}: Compared with RSMamba, this paper adopts the cross-scanning mechanism to effectively model the global context information of the image, and the computational complexity is still linear. It mainly emphasizes spatio-temporal feature interaction. 

8) \textbf{CDMamba}~\cite{10902569}: In this paper, the local feature extraction ability of convolution is explicitly fused in the Mamba architecture to enhance the detail perception ability in remote sensing binary change detection, and the bi-temporal guided fusion mechanism is used to further improve the discrimination of change regions.

9) \textbf{STRobustNet}~\cite{10879578}: The core of this thesis is to solve the problems of spatio-temporal inconsistency and feature confusion in traditional methods by using spatio-temporal robust representation, while maintaining low computational complexity.
\begin{table*}[!h]
	\centering
	\caption{Quantitative Comparisons on Three Datasets. The Best in \textcolor{red}{\textbf{Bold Red}} and The Second-Best on \textcolor{blue}{\textbf{Bold Blue}}. All Indicators are Described as Percentages.}
	\label{tab:quantitative_comparison}
	\scriptsize
	\footnotesize
	\renewcommand{\arraystretch}{1.1}
	\begin{tabular}{c|c|>{\columncolor{red!5}}c>{\columncolor{red!5}}c>{\columncolor{red!5}}c>{\columncolor{orange!10}}c>{\columncolor{orange!10}}c>{\columncolor{orange!10}}c>{\columncolor{green!10}}c>{\columncolor{green!10}}c>{\columncolor{green!10}}c>{\columncolor{yellow!15}}c}
		\toprule
		\multicolumn{2}{c|}{Method}
		& FC-EF & SNUNet & BIT & ChangeFormer & Paformer & STRobustNet & RSMamba & ChangeMamba & CDMamba & \textbf{Ours} \\
		\midrule
		&	P    & 91.17 & 89.93 & 91.55 & 90.65 & \textcolor{blue}{\textbf{91.76}} & 90.94 & 91.03 & 90.55 & 91.43 & \textcolor{red}{\textbf{92.30}} \\
		&	R    & 86.59 & 87.45 & 90.05 & 87.20 & 87.59 & 89.98 & 88.63 & 89.10 & \textcolor{blue}{\textbf{90.08}} & \textcolor{red}{\textbf{90.92}} \\
		LEVIR-CD	     &	OA   & 98.89 & 98.86 & 98.89 & 98.89 & 98.97 & 99.03 & 98.96 & \textcolor{blue}{\textbf{98.97}} & 99.06 & \textcolor{red}{\textbf{99.15}} \\
		&	F1   & 88.82 & 88.67 & \textcolor{blue}{\textbf{90.79}} & 88.89 & 89.63 & 90.45 & 89.67 & 89.82 & 90.75 & \textcolor{red}{\textbf{91.61}} \\
		&	IoU  & 79.89 & 79.65 & \textcolor{blue}{\textbf{83.14}} & 81.84 & 81.21 & 82.57 & 81.28 & 81.52 & 83.07 & \textcolor{red}{\textbf{84.51}} \\	
		\midrule
		&	P    & 92.02 & 83.34 & 91.57 & 91.85 & 92.86 & 91.85 & \textcolor{blue}{\textbf{95.68}} & 93.14 & 95.58 & \textcolor{red}{\textbf{96.32}} \\
		&	R    & 87.99 & 90.88 & 92.55 & 84.11 & 90.85 & 92.06 & \textcolor{red}{\textbf{95.50}} & 91.62 & 92.01 & \textcolor{blue}{\textbf{94.35}} \\
		WHU-CD  	     &  OA   & 97.68 & 98.91 & 99.22 & 99.07 & 97.98 & 99.36 & 98.96 & 99.40 & \textcolor{blue}{\textbf{99.51}} & \textcolor{red}{\textbf{99.66}} \\
		&	F1   & 89.99 & 86.95 & 92.06 & 87.81 & 91.85 & 91.96 & 91.56 & 92.37 & \textcolor{blue}{\textbf{93.76}} & \textcolor{red}{\textbf{95.32}} \\
		&	IoU  & 81.75 & 76.91 & 85.30 & 78.27 & 84.92 & 85.12 &  \textcolor{red}{\textbf{95.59}} & 85.83 & 88.26 & \textcolor{blue}{\textbf{90.63}} \\		
		\midrule
		&  P    & 75.93 &  \textcolor{red}{\textbf{88.43}} & 79.04 & 80.04 & 82.26 & 77.79 & 76.89 & 80.04 & 81.92 & \textcolor{blue}{\textbf{83.32}} \\
		&  R    & \textcolor{blue}{\textbf{81.03}} & 61.43 & 75.96 & 77.06 & 73.84 & 79.92 & 78.04 & 77.06 & 76.58 & \textcolor{red}{\textbf{82.16}} \\
		SYSU-CD          &  OA   & 89.47 & 89.05 & 89.58 & 90.06 & 90.08 & \textcolor{blue}{\textbf{90.72}} & 89.29 & 90.06 & 90.50 & \textcolor{red}{\textbf{91.91}} \\
		&  F1   & 78.40 & 72.64 & 77.47 & 78.52 & 77.82 & 78.84 & 77.46 & 78.52 & \textcolor{blue}{\textbf{79.16}} & \textcolor{red}{\textbf{82.74}} \\
		&  IoU  & 64.47 & 57.04 & 63.23 & 64.64 & 63.70 & 65.07 & 63.21 & 64.64 & \textcolor{blue}{\textbf{65.51}} & \textcolor{red}{\textbf{70.56}} \\
		\bottomrule
	\end{tabular}
\end{table*} 
\begin{figure*}[!t]
	\centering
	\includegraphics[width=1\linewidth]{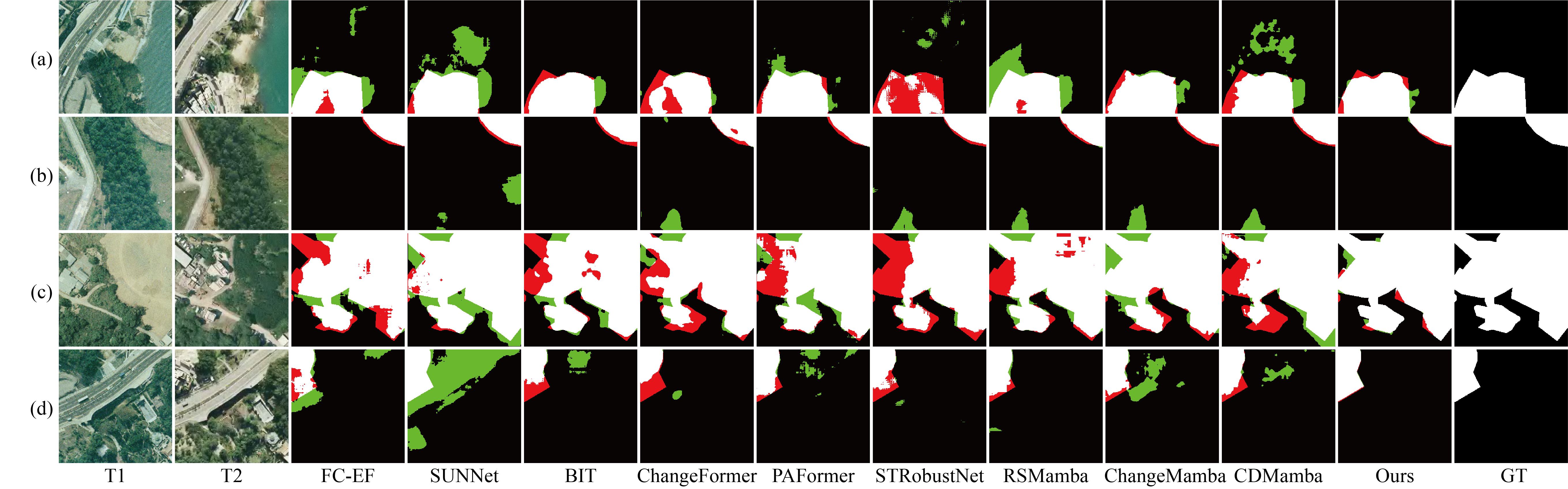} 
	\caption{Visual results of different RSCD methods on the SYSU-CD dataset. (a)–(d) are four representative samples, where white and black represent changed and unchanged areas, while red and green indicate false detection pixels and missed detection pixels.}
	\label{fig:7}
\end{figure*}
\subsection{Performance Comparison and Results Analysis}
1) \textbf{Qualitative Assessment}: To validate the effectiveness of the proposed HA2F, detailed visual BCD results across different datasets with various scenarios and scales are illustrated in Figs. \ref{fig:5} to \ref{fig:7}. The visualizations employ a color-coding scheme to distinguish the veracity of the outcomes, categorizing them into TP(signified by white), TN (rendered in black), FP (highlighted in red), and FN (denoted by green).

In RSCD tasks, large-scale changes in the scale of complex building structures, such as the transformation from farmland to urban construction land, easily lead to multi-scale difference feature matching deviation. As illustrated in Fig. \ref{fig:5}(c) and Fig. \ref{fig:6}(a), compared to other methods, HA2F is a method that exhibits superior discrimination when large-scale land grade changes conflict with subtle structural changes, even under occlusion and light interference. Moreover, as shown in Fig. \ref{fig:5}(c), HA2F shows false detection when maintaining the texture integrity of buildings and suppressing the change of vegetation land caused by seasonal differences, which fully proves its robustness in dealing with various disturbance factors.

2) \textbf{Quantitative Evaluation}: In Table \ref{tab:quantitative_comparison} the comparative evaluation of the various algorithms on the LEVIR-CD, WHU-CD and SYSU-CD datasets is summarized. This rigorous evaluation framework covers key metrics such as P, R, OA, F1, and IoU to comprehensively evaluate the performance of different algorithms.

\begin{table*}[!t]
	\centering
	\caption{Ablation study of proposed modules on WHU-CD and SYSU-CD datasets.}
	\label{tab:Ablation Studies}
	\scriptsize
	\footnotesize
	\renewcommand{\arraystretch}{1.1}
	\setlength{\tabcolsep}{10.5pt}{
		\begin{tabular}{ccc|ccccc|cccccc}	
			\toprule
			\multicolumn{3}{c|}{Models} & \multicolumn{5}{c|}{WHU-CD} & \multicolumn{5}{c}{SYSU-CD} \\
			HAFS & SAT & DFSM & P & R & OA & F1 & IoU & P & R & OA & F1 & IoU \\
			\midrule
			\ding{55} & \ding{55} &  \ding{55} & 95.56 & 92.96 & 99.44 & 94.24 & 89.11 & 85.38 & 78.75 & 91.57 & 81.93 & 69.39 \\
			
			\ding{51} & \ding{55} &  \ding{55} & 95.61 & 93.49 & 99.61 & 94.54 & 90.14 & 83.46 & 81.29 & 91.73 & 82.36 & 70.01 \\
			
			\ding{55} & \ding{51} &  \ding{55} & 95.71 & 93.54 & 99.56 & 94.61 & 89.76 & 82.67 & 81.97 & 91.58 & 82.32 & 70.27 \\
			
			\ding{55} & \ding{55} &  \ding{51} & 95.62 & 94.06 & 99.59 & 94.83 & 90.18 & 84.73 & 80.62 & 92.00 & 82.62 & 70.39 \\
			\rowcolor{green!10}
			\ding{51} & \ding{51} &  \ding{55} & \textbf{96.50} & 92.67 & 99.58 & 94.55 & 89.69 & 85.51 & 79.52 & 91.99 & 82.41 & 70.08 \\
			\rowcolor{green!10}
			\ding{51} & \ding{55} &  \ding{51} & 95.68 & 94.01 & 99.59 & 94.84 & 90.18 & \textbf{85.88} & 79.42 & \textbf{92.07} & 82.52 & 70.25 \\
			
			\ding{55} & \ding{51} &  \ding{51} & 95.64 & 93.37 & 99.57 & 94.49 & 89.55 & 83.62 & 81.58 & 91.89 & 82.59 & 70.34 \\
			\rowcolor{yellow!15}
			\ding{51} & \ding{51} &  \ding{51} & 96.32 & \textbf{94.35} & \textbf{99.66} & \textbf{95.32} & \textbf{90.63} & 83.32 & \textbf{82.16} & 91.91 & \textbf{82.74}& \textbf{70.56} \\
			\bottomrule
		\end{tabular}
	}	
\end{table*} 
\begin{figure*}[!h]
	\centering
	\includegraphics[width=1\linewidth]{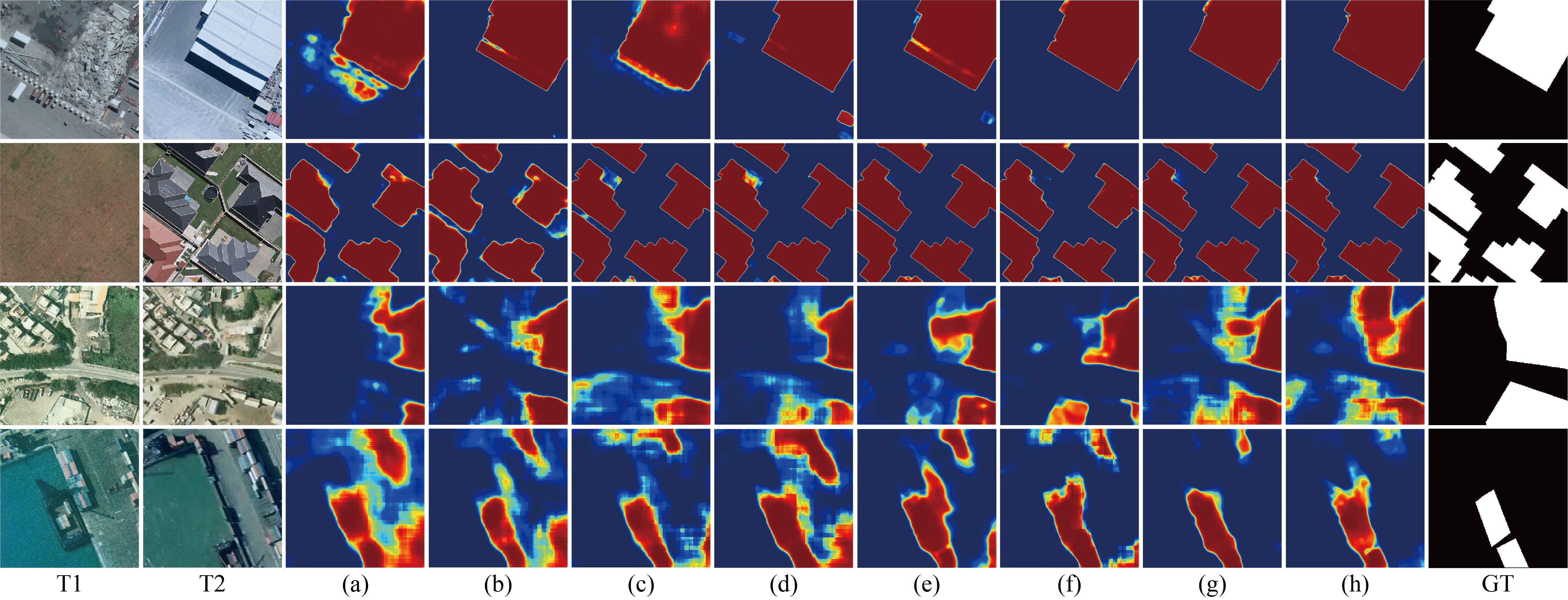} 
	\caption{Heat map visualization of different ablation modes on WHU-CD and SYSU-CD datasets. From left to right, T1, T2, (a) baseline model, (b) only HAFS, (c) only SAT, (d) only DFSM, (e) HAFS with SAT, (f) HAFS with DFSM, (g) SAT with DFSM, (f) HA2F, GT. The first two rows are the WHU-CD dataset and the last two rows are the SYSU-CD dataset.}
	\label{fig:10}
\end{figure*}

On the LEVIR-CD dataset, HA2F outperforms the latest STRobustNet and CDMamba. It achieves a significant increase in F1 and IoU, reaching 91.61\% and 84.51\%, respectively. This is because in the process of bi-temporal multi-scale feature fusion, the model can alleviate the problem of information redundancy caused by simple splicing or addition. At the same time, in the process of feature aggregation, the mutual enhancement of two-stream information is realized to improve the representation ability of features. In addition, HA2F outperforms the second-ranked BIT by 0.82\% and 1.37\% in terms of F1 and IoU metrics. This is because BIT relies on global token compression leading to detail loss, while HA2F enhances the interaction between multi-scale features and high-level features, which is more in line with the characteristics of multi-scale objects such as buildings in remote sensing images. On the WHU-CD dataset, HA2F achieves 95.32\% and 99.66\% in terms of F1 and OA, which are 3.76\% and 0.7\% higher than the second-ranked RSMamba, respectively. Compared with RS-Mamba, which is difficult to accurately locate the real semantic changes, HA2F improves the judgment accuracy of the change area through the refined feature alignment, fusion and filtering mechanism. On the SYSU-CD dataset, HA2F achieves 82.16\%, 91.91\%, 82.74\%, and 70.56\% in terms of R, OA, F1, and IoU, which are 5.58\%, 1.41\%, 3.58\%, and 5.05\% higher than CDMamba, respectively. This may be because DHFCM can fully aggregate multi-scale difference features and reduce the loss of structural information. However, NAFRM can effectively suppress false changes and is more sensitive to true changes while being more robust to irrelevant noise. In conclusion, the quantitative assessments conducted conclusively validate the effectiveness and preeminence of HA2F. The significant improvements across all evaluated metrics and datasets provide compelling evidence of the advantages of HA2F, making it a powerful tool for coping with complex RSCD tasks.

\subsection{Ablation Studies} 
To independently evaluate the contribution of each component in our proposed network, we performed ablation experiments on the HA2F model. The first row of Table \ref{tab:Ablation Studies} is the baseline model, which is a hybrid network with CNN and Vision Transformer as the backbone. We validate the effectiveness of each module by incorporating our proposed modules into the baseline model. In this section, we use the WHU-CD and SYSU-CD datasets to evaluate our ablated models, and the evaluation metrics are the same as in the comparison experiments. To maintain consistency in the experimental setup, all networks involved in the ablation studies adhere to the same implementation details as described in Section \ref{sec:Implementation Details and Evaluation Metrics}.

Table \ref{tab:Ablation Studies} shows the results of ablation experiments for this module. As can be seen from Table \ref{tab:Ablation Studies}, when the baseline model is equipped with the proposed HAFS, there is a significant improvement in both the F1 score and IoU metric. The HAFS interactively corrects the alignment bias with a unique bidirectional convolution operation mechanism, promoting the consistency of bi-temporal features outside the true change region. As a result, in the ablation experiments, the baseline model with temporal attention outperforms the baseline, particularly in terms of IoU, with improvements of 1.03 and 0.62 points on the WHU-CD and SYSU-CD datasets, respectively. Regarding the SAT, it focuses on estimating displacement vectors of pixels between adjacent frames to achieve pixel-wise non-rigid alignment. This allows SAT to compensate for geometric differences between images in the presence of viewpoint differences. The results of the ablation experiments in Table \ref{tab:Ablation Studies} also validate this observation. When the baseline model is augmented with SAT, there is a noticeable improvement in IoU, with increases of 0.65 and 0.88 points on the WHU-CD and SYSU-CD, respectively. DFSM uses the double screening mechanism of channel saliency and spatial consistency to suppress background noise and achieve more accurate change boundary detection. In the ablation experiments, when the baseline model is enhanced with DFSM, the F1 is improved by 0.59 and 0.69 points, and the IoU score is improved by 1.07 and 1.00 points on the WHU-CD and SYSU-CD datasets, respectively. Conversely, when DFSM is removed from our final model, there is a decrease in metrics, with a decrease of 0.77 and 0.33 points in the F1 score on the two datasets, respectively. Fig. \ref{fig:10} presents a heatmap visualization of the module ablation experiment. From the visual results (a), (b) and (g) proves that HAFS can reduce the loss of structural information and has a positive impact on dual-temporal feature fusion. Furthermore, the visualizations in (a), (c), and (f) reveal that SAT can effectively highlight edge difference information and strengthen edge information. Additionally, the results in (a), (d), and (e) demonstrate DFSM effectively filters noise information and reduces spurious changes to a certain extent. In summary, the ablation heatmap comparison validates the effectiveness of the proposed module. 

\begin{figure*}[!t]
	\centering
	\includegraphics[width=0.9\linewidth]{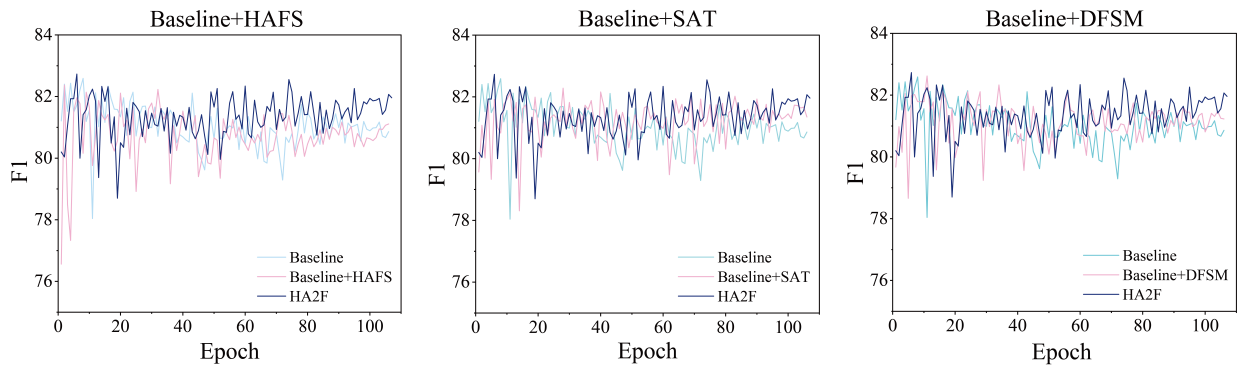} 
	\caption{The curve of F1 score on the SYSU-CD training set. From left to right: Baseline+HAFS, Baseline+SAT, and Baseline+DFSM.}
	\label{fig:11}
\end{figure*}

Furthermore, Fig. \ref{fig:11} shows the variation curve of F1 in the training set. In the change detection task, since the changed region is usually much smaller than the unchanged region, there is a significant imbalance in the sample distribution, which may cause the model to deviate from the learning direction in some training batches, resulting in indicator fluctuations. To this end, we evaluate the model performance on the validation set every few rounds during the training process, and record F1, IoU and other indicators. Finally, we select the model with the highest F1/IoU performance and relatively stable on the validation set to avoid overfitting and ensure the reliability of the results. It can be seen from the figure that the F1 values of the three modules are higher than that of the baseline model as a whole and show a trend of convergence towards stability, indicating that they all have a positive effect on RSCD. In addition, the F1 value of HA2F is always maintained at a high level, which further verifies the effectiveness and superiority of the module.

\begin{table}[!t]
	\centering
	\caption{Quantitative comparison of different convolutional layers in HAFS on the SYSU-CD dataset.}
	\label{tab:Ablation Studies2}
	\scriptsize
	\footnotesize
	\renewcommand{\arraystretch}{1.1}
	\setlength{\tabcolsep}{8pt}{
		\begin{tabular}{c|c|ccccc}	
			\toprule
			\multirow{2}{*}{HAFS} &\multirow{2}{*}{Variants} & \multicolumn{5}{c}{SYSU-CD} \\
			&&  P & R & OA & F1 & IoU   \\
			\midrule 
			\multirow{3}{*}{CBR}  & 3L-Conv & 81.07 & 79.15 & 90.41 & 80.10 & 66.89 \\
			& 2L-Conv & 81.11 & 79.23 & 90.46 & 80.20 & 67.27 \\
			& \cellcolor{yellow!15}1L-Conv & \cellcolor{yellow!15}\textbf{81.46} & \cellcolor{yellow!15}\textbf{80.09} & \cellcolor{yellow!15}\textbf{90.52} & \cellcolor{yellow!15}\textbf{80.76} & \cellcolor{yellow!15}\textbf{68.01} \\
			\midrule 
			\multirow{3}{*}{DBR}  & 3L-DSC & 81.71 & 78.64 & 90.41 & 80.14 & 66.42 \\
			& \cellcolor{green!10}2L-DSC & \cellcolor{green!10}\textbf{82.14} & \cellcolor{green!10}79.45 & \cellcolor{green!10}90.49 & \cellcolor{green!10}80.77 & \cellcolor{green!10}67.35 \\
			& \cellcolor{yellow!15}1L-DSC & \cellcolor{yellow!15}82.06 & \cellcolor{yellow!15}\textbf{80.26} & \cellcolor{yellow!15}\textbf{90.69} & \cellcolor{yellow!15}\textbf{81.15} & \cellcolor{yellow!15}\textbf{68.21}\\
			\bottomrule
		\end{tabular}
	}
\end{table} 

\begin{table}[!t]
	\centering
	\caption{Comparison of different methods are replaced in SAT.}
	\label{tab:Ablation Studies4}
	\scriptsize
	\footnotesize
	\renewcommand{\arraystretch}{1.1}
	\setlength{\tabcolsep}{10.5pt}{
		\begin{tabular}{c|ccccc}	
			\toprule
			\multirow{2}{*}{Methods} & \multicolumn{5}{c}{SYSU-CD} \\
			& P & R & OA & F1 & IoU \\
			\midrule
			DCN & 82.11 & 80.04 & 91.52 & 81.06 & 68.24  \\ 
			DyConv   & 81.42 & 80.57 & 91.50 & 80.99 & 68.56 \\
			CrossAttn  & 80.98 & 79.61 & 91.47 & 80.29 & 67.74\\
			\rowcolor{yellow!15}
			SAT  & \textbf{82.67} & \textbf{81.97} & \textbf{91.58} & \textbf{82.32} & \textbf{70.27} \\
			\bottomrule
		\end{tabular}
	}
\end{table}

\begin{table}[!t]
	\centering
	\caption{Ablation study of DFSM on SYSU-CD datasets.}
	\label{tab:Ablation Studies3}
	\scriptsize
	\footnotesize
	\renewcommand{\arraystretch}{1.1}
	\setlength{\tabcolsep}{9.5pt}{
		\begin{tabular}{cc|ccccc}	
			\toprule
			\multicolumn{2}{c|}{DFSM} & \multicolumn{5}{c}{SYSU-CD}  \\
			SCF&CSF &  P & R & OA & F1 & IoU  \\
			\midrule 
			\rowcolor{green!10}
			\ding{55} & \ding{55}  & \textbf{85.38} & 78.75 & 91.57 & 81.93 & 69.39 \\
			\ding{51} & \ding{55}  & 83.74 & 80.24 & 91.55 & 81.96 & 68.72 \\
			\ding{55} & \ding{51}  & 83.66 & 80.19 & 91.54 & 81.89 & 68.56 \\
			\rowcolor{yellow!15}
			\ding{51} & \ding{51}  & 84.73 & \textbf{80.62} & \textbf{91.64} & \textbf{82.62} & \textbf{70.39} \\
			\bottomrule
		\end{tabular}
	}
\end{table}
 
In addition, more detailed ablation experiments are carried out for the three modules. Table \ref{tab:Ablation Studies2} shows the ablation study of different convolutional layers in HAFS on SYSU-CD dataset. It can be seen that in both CBR and DBR, the F1 and IoU indicators of one layer convolution are the highest, which are 80.76\%, 68.01\% and 81.15\%, 68.21\% respectively. It is possible that the single-layer convolution module can effectively prevent overfitting and improve generalization ability due to its simple structure. Table \ref{tab:Ablation Studies4} shows the ablation results for SAT. Replace it with three different methods: DCN~\cite{wang2023internimage}, DyConv~\cite{10924169} and CrossAttn~\cite{10889098}. It can be seen that all the indicators of SAT are the highest, among which F1 and IoU are 82.32\% and 70.27\% respectively. This shows that SAT can achieve precise spatial alignment compared with other methods. As can be seen from the Table \ref{tab:Ablation Studies3}, DFSM mainly includes two parts: Spatial Consistency Filter (SCF) and Channel Saliency Filter (CSF). When the module has only SCF or CSF respectively, none of the indicators except the R indicator is improved. It indicates that a single filter cannot effectively suppress noise. When two filters are present at the same time, both F1 and IoU reach their highest values, which are 82.36\% and 70.01\%, respectively. This indicates that only by combining SCF and CSF, DFSM can fully suppress the noise interference caused by shadows in irrelevant regions. 


\begin{table*}[!t]
	\centering
	\caption{Ablation evaluations of different methods are replaced in DHFCM and NAFRM of the framework, respectively.}
	\label{tab:Ablation DHFCM and NAFRM Studies}
	\scriptsize
	\footnotesize
	\renewcommand{\arraystretch}{1.1}
	\setlength{\tabcolsep}{12.5pt}
	\begin{tabular}{c|ccccc|ccccc}	
		\toprule
		\multirow{2}{*}{Methods} & \multicolumn{5}{c|}{WHU-CD} & \multicolumn{5}{c}{SYSU-CD} \\
		& P & R & OA & F1 & IoU & P & R & OA & F1 & IoU\\
		\midrule
		\rowcolor{green!10}
		3D-DEM & 95.15 & 93.04 & 99.55 & 94.08 & 89.67 & \textbf{84.31} & 79.75 & \textbf{91.74} & 81.96 & 69.18 \\ 
		MSAA   & 95.04 & 93.01 & 99.49 & 94.01 & 88.65 & 82.94 & 81.13 & 91.62 & 82.03 & 69.24 \\
		\rowcolor{green!10}
		FEM    & 94.98 & 92.91 & 99.30 & 93.93 & 89.26 & 80.66 & \textbf{81.92} & 91.48 & 81.28 & 68.90 \\
		\rowcolor{yellow!15}
		\textbf{DHFCM | Ours}  & \textbf{95.61} & \textbf{93.49} & \textbf{99.61} & \textbf{94.54} & \textbf{90.14} & 83.46 & 81.29 & 91.73 & \textbf{82.36} & \textbf{70.01} \\ \bottomrule
		\rowcolor{green!10}
		GA     & 94.98 & 91.18 & 99.44 & 93.04 & 88.05 & 82.18 & \textbf{82.11} & 91.58 & 82.15 & 69.70 \\ 
		EUCB   & 95.28 & 92.91 & 99.51 & 94.08 & 88.73 & 85.44 & 78.32 & 91.49 & 81.73 & 69.35 \\
		\rowcolor{green!10}
		MASAG  & \textbf{96.16} & 92.26 & 99.52 & 94.17 & 89.01 & \textbf{85.78} & 78.84 & \textbf{91.93} & 82.16 & 69.73 \\
		\rowcolor{yellow!15}
		\textbf{NAFRM | Ours}  & 95.64 & \textbf{93.37} & \textbf{99.57} & \textbf{94.49} & \textbf{89.55} & 83.62 & 81.58 & 91.89 & \textbf{82.59} & \textbf{70.34} \\
		\bottomrule
	\end{tabular}
\end{table*} 


\begin{figure}[!t]
	\centering
	\includegraphics[width=1\linewidth]{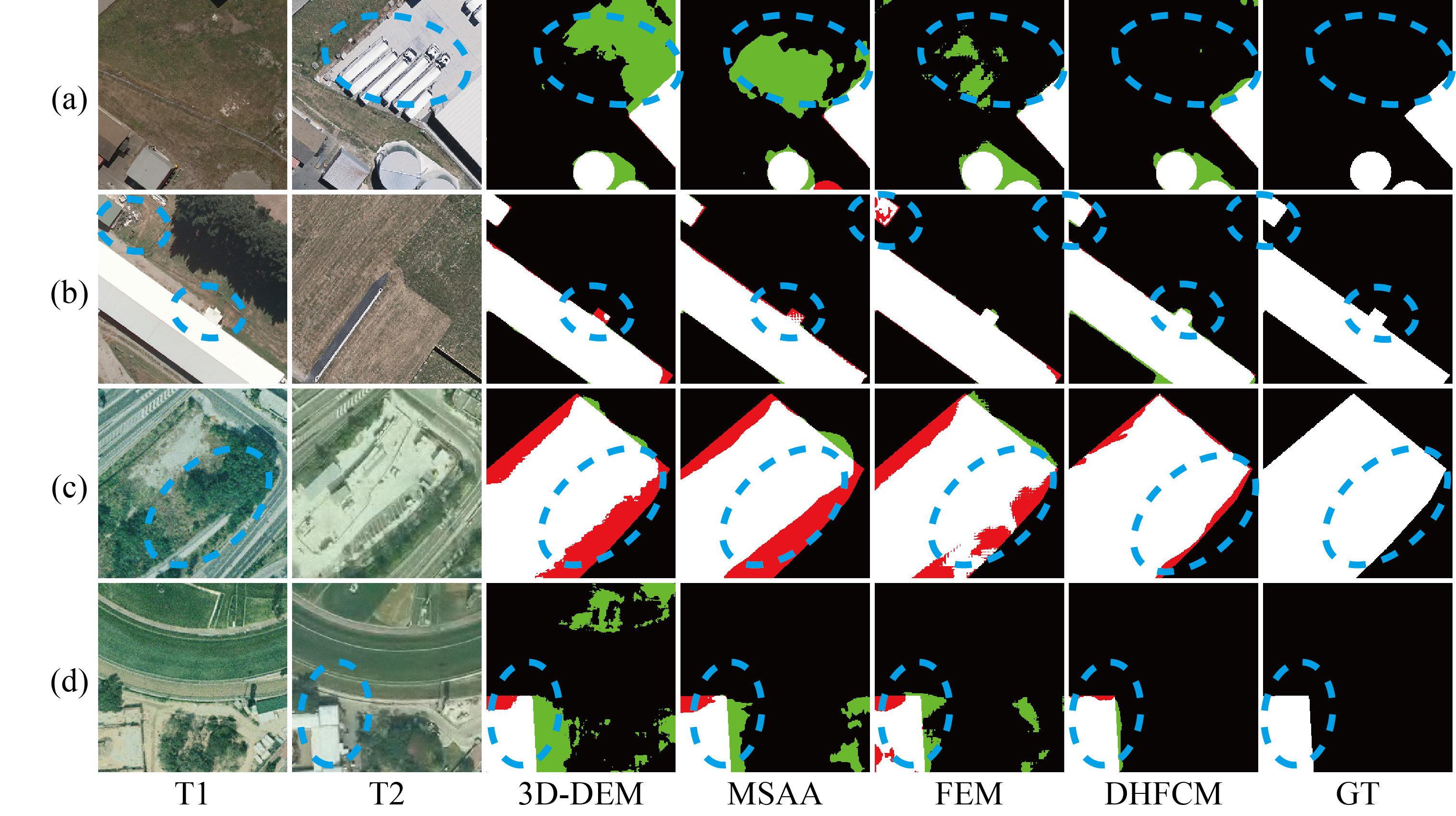} 
	\caption{Visual results of some different methods for DHFCM. (a)–(b) are on the WHU-CD dataset and (c)–(d) are on the SYSU-CD dataset. White and black represent changed and unchanged areas, while red and green indicate false detection pixels and missed detection pixels.}
	\label{fig:8}
\end{figure}
Moreover, we also analyze the ablation experiments of the two designed modules, which are DHFCM and NAFRM, compared with other methods. Table \ref{tab:Ablation DHFCM and NAFRM Studies} shows the results under different methods. DHFCM is replaced by 3D-DEM~\cite{liu2025edge}, MSAA~\cite{liu2024cm}, and FEM~\cite{10423050}. NAFRM is replaced by GA~\cite{wei2024spatio}, EUCB~\cite{rahman2024emcad}, and MASAG~\cite{kolahi2024msa}. We can observe from Table \ref{tab:Ablation DHFCM and NAFRM Studies} that different methods perform well on the training samples, but DHFCM outperforms the others. There is little difference in recall, but an improvement in both F1 and Iou. This indicates that DHFCM plays a role in adjusting the bias of multi-scale fusion. The results on WHU-CD and SYSU-CD datasets are 94.54\%, 90.14\% and 82.36\%, 70.01\%, respectively. Table \ref{tab:Ablation DHFCM and NAFRM Studies} shows that the different methods basically perform worse than NAFRM in the decoder stage. Although there is little difference in precision, there is an improvement in recall. This indicates that NAFRM effectively captures change signals with fuzzy boundaries and suppresses spurious changes. In order to pursue high accuracy, the comparison method is too conservative and filters out the true changes that are difficult to judge. Therefore, NAFRM is able to achieve the best F1 score compared to other methods, with results of 94.49\% and 82.59\% on WHU-CD and SYSU-CD datasets, respectively. 

\begin{figure}[!t]
	\centering
	\includegraphics[width=1\linewidth]{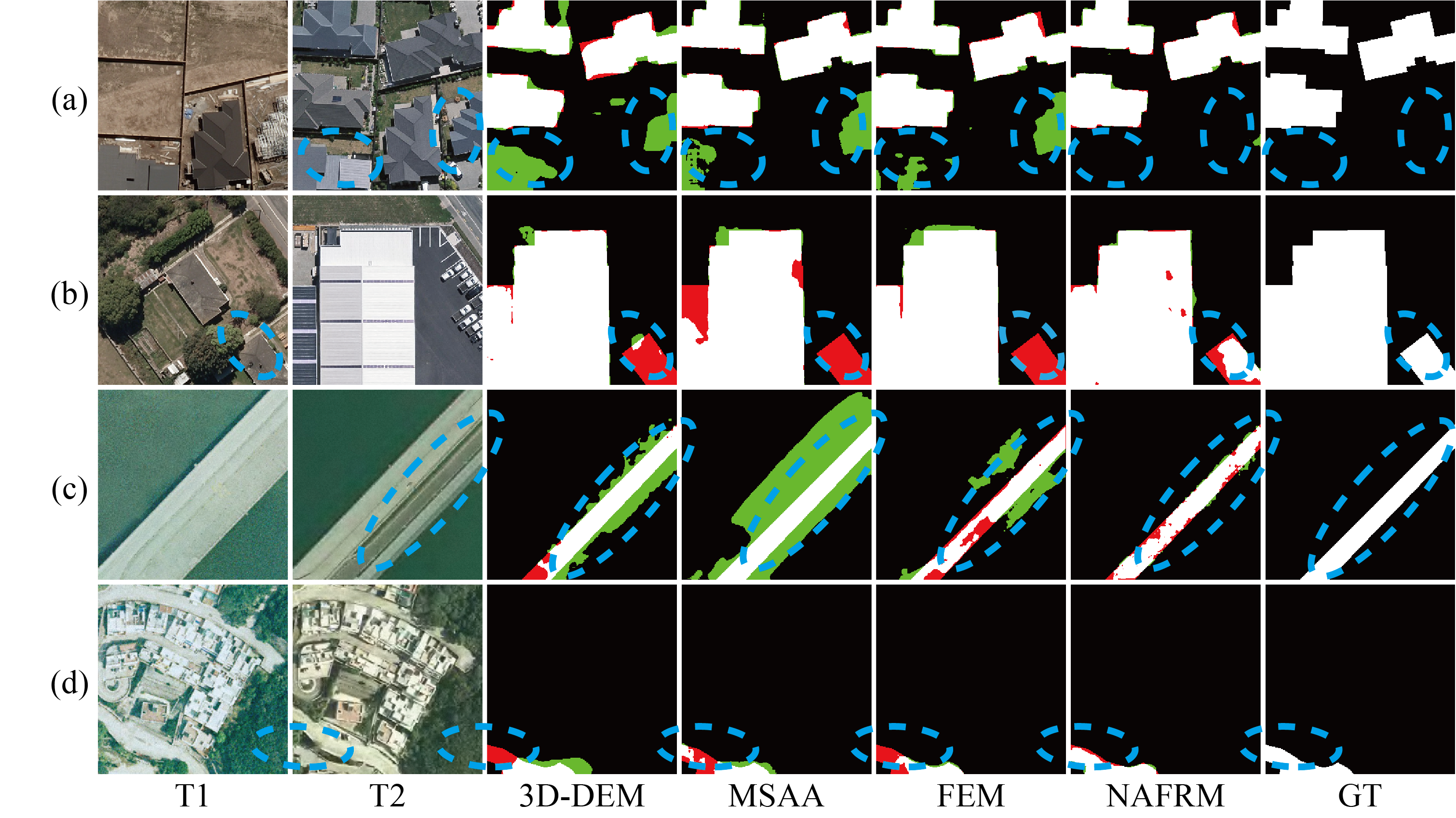} 
	\caption{Visual results of some different methods for NAFRM. (a)–(b) are on the WHU-CD dataset and (c)–(d) are on the SYSU-CD dataset. White and black represent changed and unchanged areas, while red and green indicate false detection pixels and missed detection pixels.}
	\label{fig:9}
\end{figure}
To verify the effectiveness of the proposed modules, the Fig. \ref{fig:8} and \ref{fig:9} provide a visualization of the module ablation experiments. From the Fig. \ref{fig:8} (a), (b) and (c), it is evident that DHFCM has a positive impact on both large-scale scenes and small target changes in complex background environments. Furthermore, the visualizations in Fig. \ref{fig:9} (a), (c), and (d) reveal that NAFRM reduces the spurious variation to some extent while preserving the regions of actual variation. In conclusion, the ablation experiment verified the effectiveness of the proposed module.

\section{CONCLUSION}
\label{CONCLUSION}
In this study, we propose a hybrid CNN-Transformer change detection network HA2F. The network mainly includes two key modules: DHFCM and NAFRM. DHFCM enhances the obtained multi-scale information by establishing the dynamic dependence between adjacent levels of features, and effectively integrates the complementary advantages between different scales. In addition, in order to suppress the spurious variation caused by noise, the decoder integrates NAFRM to improve the resistance to noise interference and improve the robustness. Extensive experiments on three benchmark datasets show that the proposed HA2F outperforms other state-of-the-art methods in terms of both effectiveness and efficiency. In the future, future work should focus on exploring lightweight network architectures and developing an advanced framework for integrated semi-supervised learning to alleviate the dependence of model training on large-scale labeled datasets.

\bibliographystyle {IEEEtran}
\bibliography {reference}
\end{document}